# Integrating Imitation Learning with Human Driving Data into Reinforcement Learning to Improve Training Efficiency for Autonomous Driving

Heidi Lu
The Harker School, 500 Saratoga Avenue
San Jose, CA 95129

Abstract

Two current methods used to train autonomous cars are reinforcement learning and imitation learning. This research develops a new learning methodology and systematic approach in both a simulated and a smaller real world environment by integrating supervised imitation learning into reinforcement learning to make the RL training data collection process more effective and efficient. By combining the two methods, the proposed research successfully leverages the advantages of both RL and IL methods. First, a real mini-scale robot car was assembled and trained on a 6 feet by 9 feet real world track using imitation learning. During the process, a handle controller was used to control the mini-scale robot car to drive on the track by imitating a human expert driver and manually recorded the actions using Microsoft AirSim's API. 331 accurate human-like reward training samples were able to be generated and collected. Then, an agent was trained in the Microsoft AirSim simulator using reinforcement learning for 6 hours with the initial 331 reward data inputted from imitation learning training. After a 6-hour training period, the mini-scale robot car was able to successfully drive full laps around the 6 feet by 9 feet track autonomously while the mini-scale robot car was unable to complete one full lap round the track even after 30 hour training pure RL training. With 80% less training time, the new methodology produced significantly more average rewards per hour. Thus, the new methodology was able to save a significant amount of training time and can be used to accelerate the adoption of RL in autonomous driving, which would help produce more efficient and better results in the long run when applied to real life scenarios.

I. Introduction:

WHO reported that 1.3 million people die every year in the US due to traffic-related accidents, and nearly 3,500 lives per day can be saved [1]. According to Accenture and the Stevens Institute of Technology in Hoboken, N.J., the insurance industry, worth more than $225 billion, will see traditional premiums drop by nearly 20% over the next 30 years [3]. Autonomous driving is quickly becoming a reality. In the imitation learning process, the agent observes and learns from expert demonstrations [4]. However, collecting sufficient diverse data is challenging because samples tend to be positively biased

and the agent's performance is limited by the expert driver's performance. In the reinforcement learning process, the agent learns from trial and error and a designated reward function, but data collection is incommodious as focusing on only positive scenarios leads to biased sampling. Physical damage to the car is bad as sampling data can be extremely costly [5].

Most autonomous driving cases have been solved so far and many cars that are testing and driving around nowadays can be seen with a safety driver behind the wheel. Most autonomous driving companies are currently focusing on solving the last 10% corner cases in order for them to remove the safety driver (ensuring 99.99% safety) and start making business and profits. Waymo has removed the safety driver in a very limited region in Phoenix in its robot taxi service, but not all areas [10]. Supervised machine learning (ML) is commonly used now by autonomous driving companies, but requires heavy data labelling, and is not a good solution for solving the last 10% corner cases. It only knows how to deal with labelled scenarios, but normally corner cases are the ones non-experienced and not labelled previously. RL is a different approach as it learns from its own trial-and-error, and thus doesn't require human labelling data [8]. It can solve unseen corner cases from similar situations it's experienced before, similar to human drivers. Unfortunately, collecting diverse driving data for RL is very challenging, and that's why RL is very rarely used by most autonomous driving companies. Current RL development and methods are still primarily on research papers[13]. However, in the future, RL will be implemented by autonomous driving companies by combining supervised ML in their software, focusing on the areas of motion planning, which will help solve the complicated corner cases. Other normal driving cases can be solved by rule based coding and supervised ML through labelled data.

This paper proposes a new methodology which integrates imitation learning with efficient human driving data to accelerate reinforcement learning for autonomous driving with a more efficient and effective training data collection process. By combining both methods, the advantages of each can be leveraged to enhance the performance of reinforcement learning from utilizing guidance from a human driver. Imitation learning serves as a foundation to bootstrap reinforcement learning training by providing the necessary initial training data and a basic pretrained model for reinforcement learning. Ultimately, reinforcement learning does not limit the agent's performance and requires no expensive data labelling cost, which are major advantages over imitation learning, and thus can be used to effectively solve the last 10% autonomous driving corner cases.

II.  Related Works

As can be seen from Table 1, most of the previous research primarily focuses on proving that RL is more effective than supervised IL, or that combining supervised IL with RL is better than just pure IL or pure RL in terms of performance. With the development of deep reinforcement learning, the domain of reinforcement learning has become a powerful learning framework now capable of learning complex policies in high dimensional environments [18]. A group of researchers wrote a research paper in 2019 that summarizes deep reinforcement learning (DRL), provides a taxonomy of automated driving tasks where (D)RL methods have been employed, highlights key challenges and methods to evaluate, test, and robustify existing solutions in RL and IL. They formalized and organized RL applications for autonomous driving and found that it suits RL but still suffers from many challenges that need to be resolved. Past researchers Tianqi Wang et al. [5] in October of 2019 sought to combine imitation learning and reinforcement learning in a simulated environment to train the car to drive in different weather environments. Following this, also in October of 2019, researchers Fenjiro Youssef et al. [17] wrote a paper on the optimal combination of IL and RL for self-driving cars. Testing it in a simulated environment, their resulting Advantage Actor-Critic algorithm from Demonstrations optimally Constrained (A2CfDoC) model was able to outperform several existing algorithms in terms of speed and accuracy as well as surpassing the expertise level using RL. Additionally, in September of 2020, Rousslan F. J. DOSSA et al. [11] conducted an experiment on the hybrid of RL and IL for human-like agents. They proposed that these hybrid agents demonstrated similar behavior to that of human experts and found that these hybrid agents were able to surpass the human expert in each given scenario.

In summary, all previous research works were trying to prove how combining these two methods are better than just one or using enhanced algorithms to test these concepts in simulated environments. However, it is also important to focus on how to apply these methods and research papers to the real world's autonomous driving works. This research work aims to fill the research gap and it is also the first research focusing on pretraining IL to effectively generate good reward datasets for bootstrapping RL, as RL is the ultimate solution to the last 10% corner cases. A mini-scale robot car has been built and used to test the proposed methodology in a smaller, real-world environment. In the test, a handle controller was used to drive the mini-scale robot car and the actions were recorded using Microsoft Airsim's API to generate accurate human-like performance for the "reward" datasets used in subsequent RL training. The test result based on robot car performance shows that the proposed methodology helped save 80% of training time as compared to pure RL, and it also produced better performances.

III.   Methodology

The complete comparison of the proposed new architecture over previous existing architecture is shown in Figure 2. Figure 2A shows a demonstration of how driving using machine learning is better than human driving and how the existing architecture focuses on proving and making the rewards generated by reinforcement learning is better than the reward data generated by human expert driving data.

Figure 2B shows a diagram of how quickly Mk can achieve a similar level to human expert driving, and how the new architecture works on making the reward generated by reinforcement learning better than the reward generated by human expert driving data in less training time.

The IL agent can be only as good as the expert's demonstrations and since it lacks generalization due to expensive labelling costs, it's not good to use pure IL. However, when facing unknown environments, RL and IL combined turn out to be the greatest strengths and can achieve the highest performance of the agent, and their capacities to manage new situations by learning through the process of trial and error and exploration makse them favorable.

The key difference between the proposed architecture shown in Figure 2B and pre-existing architecture shown in Figure 2A is that **R**Expert generated by **A**Expert was used as the inputs to accelerate the proposed **R**original generated by the reinforcement learning agent, **A**Original, while other researchers used **R**Expert generated by **A**Expert to compare to **R**original generated by the reinforcement learning agent, **A**Original. Pre-existing architecture focused on proving and making **R**original better than **R**Expert, while the proposed architecture worked on making **R**original similar to **R**Expert but achieving it in less training time. This research believes that the robot should drive as good as or better than a human, while retaining some of its high performance of RL.

$$R = \begin{cases} 0, \text{ if } R_{distance}, \text{ or } R_{speed} < 0.1 \\ W_d * R_{distance} + W_s * R_{speed} \end{cases}$$

$R_{distance} = Min(1, D/D_i)$, $R_{speed} = Min(1, V/V_i)$
$W_d = W_s = 0.5$
$D_i = 10$ cm to track edge
$V_i = 10$ cm/s
**One reward is generated when R is close to 1**

In the proposed new architecture, R is the reward function defined in Figure 2B. Here the distance to the nearest obstacle and current vehicle speed is considered as two components of the reward function to make sure that only keeping far from the obstacles and driving fast at the same time can result in a high reward. **R**Expert is the reward data generated from the handle, for example, by "expert" human driving as **A**Expert. It works as the input for reward samples in the DDPG (Deep Deterministic Policy Gradient), the popular RL algorithm used as the reinforcement learning algorithm. $R_{distance}$ is the distance to the nearest obstacle, in othisur case, the road edges, and $R_{speed}$ is the jetbot robot car's current speed. Two important reward functions → "D!" is the threshold value chosen for ideal distance (in this research, it was chosen to be 10 CM from the edge). "W!" is the threshold value chosen for ideal speed, which is 10 CM/s. Wd and Ws are the weights of reward distance and speed and sum up to one, and in this research Wd = Ws = 0.5. Based on this setting, the reward function is able to be squeezed to the inter-value [0,1], meaning that reward can only be two numbers, 0 or 1. 10 cm was chosen as the ideal distance because the jetbot robot car is centered at 10 cm away from the edge. **R**original is the actual reward generated by reinforcement learning. **A**Original is the agent used in the RL simulator.

IV. Experimental Set-up

As seen in Figure 3A and 3B, the mini-scale robot car is composed of different hardware and software parts. Combined, it is a multi-functional jetbot robot car powered by the Nvidia JETSON NANO AI platform. The Jetson Nano 64G image Nvidia firmware was burned to a 64G SD card. After assembling and building up its firmware environment, the jetbot robot car was assigned an IP and connected to wifi.

The hardware part of the vehicle is made of green aluminum alloy and follows a unique mechanical structure of a tractor. It is also equipped with 3-degrees of freedom lifting platform and a 8 million HD Raspberry Pi camera, which gives a real-time view of front looking scenes. The microprocessor is a Quad-core ARM A57 + 128-core Nvidia Maxwell. The operating system is a Ubuntu18.04 LTS with the input images through a custom HD camera and the output consisting of a L-type 370 motor, buzzer, 3-DOF camera platform, and an OLED display. The power solution is a 18650 battery pack with 12.6V [23].

*A. Setting up the track:*

As can also be seen in Figure 3C, the track used as the testing environment is a 6 feet by 9 feet automatic drive track map. It is made of durable, tarpaulin waterproof material and is a large size, replicating a real life road and track environment. This track was specifically chosen because it was most suited for the small-size front looking camera used by the jetbot robot car. It is composed of two tracks of suitable width, with the white dotted lines and the yellow solid lines on both sides able to be used as reference objects to control the movement direction of the jetbot robot car. Compared with an ordinary circular track, this track possesses some 90 degree angles, as well as different degrees of corners and a simulated sidewalk which helps achieve creative functions such as stop signs and pedestrian avoidance.

*B. Collecting the data:*

One of the most important new breakthroughs was the reward data collection process, and it used Airsim API to visualize and record images with labels for rewards. Each image generates a set of X, Y values of a reward corresponding to speed and steering angle. As seen in Figure 3D, by using the handle controller to manually drive the robot car at different locations on the track, a "green dot" was placed at the target direction in each location as a reward and it helps collect a total of 331 initial reward datasets. As seen in Figure 4, the dotted light blue lines with triangles represent the collected 331 reward datasets used for pure imitation learning.

*C. Training the model:*

A block diagram of the end-to-end RL training system from Nvidia is shown in Figure 1. Images collected for steering gear (rotation) and motor (speed) are fed into a convolutional neural network (CNN) which then computes a proposed steering command. The proposed command is compared to the desired command for that image and the CNN is adjusted to bring the CNN output closer to the desired output. The adjustment is accomplished using back propagation as implemented in Nvidia's machine learning package. Once trained, the CNN can generate steering angle and speed from the video images of the single center front camera. Steering and speed commands generated by the CNN can control the cars through a drive-by-wire interface.

For this project, the smallest network ResNet-18 was used because it provides a good balance of performance and efficiency for the Jetson Nano platform. The chosen action (speed, steering) is concatenated in the final fully connected layer's input. The neural network was trained to get an input image and output set of x (speed), y (steering) values corresponding to a target. The model training consisted of 5 main steps. 1) Using PyTorch deep learning framework to train the model to identify road conditions for autonomous driving. 2) Creating a custom "torch.utils.data.Dataset" for loading the 331 images collected and parsing the x and y values in the image file name. 3) Splitting the dataset into training (90% of the data) and testing (10%) data that was used to verify the accuracy of the model trained. 4) Training the regression model by using a data batch size of 64 based on the Jetson Nano GPU and setting the number of iterations to 50, then training the pure RL models for increasing increments of time (3 hours, 6 hours, 15 hours, 30 hours) and the new proposed model for only 6 hours. In the project, 50 times took approximately 1 hour to train. 5) Lastly, testing .pth files were generated for the trained models.

*D. Uploading the trained algorithm:*

Next, it was time to upload the trained algorithm to the jetbot robot car and implement the motion algorithm using the proportional integral differential (PID) controller to control the car. Then, values were assigned and adjusted to the following indexes through the slider to get the Jetbot robot car to drive to its best condition: A) Speed_gain_index- setting the speed value to start the car. B) Steering_gain_index- if the car is spinning, this index is reduced until the driving becomes smooth. C) Steering_bias_index- if the car leans too far to the right or too far to the left of the track, it adjusts this index until the robot car is centered on the track again.

*E. Testing and Optimization:*

Lastly is track testing of autonomous driving algorithms in the 6 feet by 9 feet driving track map. The autonomous driving testing performs as the following steps - a. processing camera images; b. performing a neural network based on a trained model; c.

calculating the approximate steering value and vehicle speed for each camera image; d. controlling the motor of the jetbot robot car using the PID controller. The testing result was measured based on the number of interventions when the car runs off the track. In the end, the proposed new methodology was able to achieve 100%, full autonomous driving after only 6 hours of training (testing results can be found in reference file 1.)

V. Results and Discussion:

In this paper, the number of rewards collected after training for a certain amount of time was used in order to compare learning efficiency for these 3 methods, pure RL, pure IL, and the newly proposed method. The results shown in Figure 4 is an important graph of the Average Accumulated Reward where the orange star represents the method, the blue triangle represents pure IL, and the black squares represents pure RL. One can see the significant difference between the slow progress of the pure RL method and the straight line of pure IL. These numbers of the reward function are important to and could heavily affect the performance of the trained agent.

The original IL's generated reward # - **331** collected rewards, while the reward policy obtained from pure RL (DDPG) which was trained from scratch generated only **11** rewards in 30 hours training, never performed well, and showed very little improving trends. The reward policy obtained from the new method achieves a considerable performance boost from the original IL's trained policy - **376** rewards in 3 hours of training and **431** rewards in 6 hours of training.

Loading the following trained models to run the autonomous driving program, 3 pure RL models were trained for 3 hours, 6 hours, and 30 hours.
1. After 3 hours, the robot car was unable to move autonomously for even 2ft.
2. After 6 hours, the robot car did not show much progress, unable to turn corners or drive autonomously for 4ft.
3. After 30 hours, the robot car was still unable to complete a half loop of the track.

However, after training the new RL model with IL rewards for only 6 hours, the robot car was able to finish multiple full track loops smoothly, using 80% less training time.

The percentage of the time the trained model can drive autonomously is determined by counting simulated human interventions. In this project, the interventions occur when the mini-scale robot car runs completely off the track. In real life an actual intervention would require a total of six seconds: this is the time required for a human to retake control of the vehicle, re-center it, and then restart the self-steering mode.

**Autonomy Value = (1 − (number of interventions) · 6 [seconds] / testing time [seconds]) * 100**

As a result, approximately 95 interventions in 600 seconds (or 10 minutes) were recorded during testing for pure RL, while 0 interventions in 600 seconds was recorded testing for the new method.

- The calculated autonomous value for pure RL is:
  (1− 95 · 6/600) ·100% = 5%
  (95% of the time is off the track)
- The autonomy value of the new method is 100%.

VI. Real World Autonomous Driving Applications

The proposed new method has significant potential when applied to real world autonomous driving development, primarily because it can be easily scaled by having 1 human driven car as an efficient agent driving around in the real roads for data collection. The experiment testing environment can be quickly extended to cover thousands of miles of road, which can help increase RL's sampling efficiency, avoid data overfitting, and thus improve versatility and generality of the RL algorithm. Therefore, the suggested methodology saves a significant amount of training time and accelerates the quick adoption of RL in autonomous driving, which would produce more efficient and better results in the long run when applied to real life scenarios.

VII. Conclusion

This research has presented and demonstrated the combination of IL and RL for the first time and showed a significant advantage in RL training efficiency with the utilization of pre-collected reward data from IL. The training results show that although training via pure RL and training via combined IL and RL would eventually lead to a similar performance, around 80% of training time was able to be cut down by initially training with the reward data collected by imitation learning. The RL method heavily relies on the computing power of the control unit, and the Nvidia Jetson Nano is a relatively small platform with only 64 Core ARM CPU and 128 Core Nvidia Maxwell GPU. If using more powerful control units than Jetson Nano, the actual saving in training time for combining imitation learning and reinforcement learning may be significantly less than 80%.

As compared to supervised IL that requires explicit data labeling, the proposed method was much simpler, a main advantage being that the experiments only needed 1 remote controlled robot car and 1 simulated agent car to learn how to handle interaction. The jetbot robot car trained using the proposed methodology seems to be able to drive better and smoother than the robot car trained using pure RL, because the driving policy is rewarded through imitating the collected human expert driving behaviors. Still, the jetbot robot car was only tested in a very simple loop track without any complicated scenarios, such as lane merging, mingling with other objects such as bikers, human driving cars, pedestrians, stop signs, traffic lights, etc. In order to solve the last 10% corner cases in autonomous driving, there's not enough "bad" examples and one can't simply hire someone to crash their cars purposely to produce the "penalized" driving data for raining the reinforcement learning model. The human expert driving helps collect only "reward" data for reinforcement learning, while the "bad or penalized" data for reinforcement learning to be generated by human expert driving can be costly and tragic. In reality, most car accidents were caused by "bad" driving behaviors which is the main challenge all research projects face. If in a really complicated driving environment with other objects, the car may not be able to drive better or smoother. As a result, the suggested methodology proves to save a significant amount of RL training time and can be quickly applied to real world autonomous driving development, but still more work is needed to verify its robustness in solving the rare complicated autonomous driving corner cases.

Reference file 1:

https://www.youtube.com/watch?v=BeFd6wfJTxI.

Table 1 Summary of proposed approach compared with previously reported research

| Research Focus | [Improved Reinforcement Learning through Imitation Learning Pre-training Towards Image-based Autonomous Driving](#) | [Optimal Combination of Imitation and Reinforcement Learning for Self-driving Cars](#) | [Hybrid of Reinforcement and Imitation Learning for Human-Like Agents](#) | Proposed Research: New Methodology To Accelerate Reinforcement Learning For Autonomous Driving Using Imitation Learning & Human Driving Data |
|---|---|---|---|---|
| Research Group | School of Electrical Engineering, Korea Advanced Institute of Science and Technology, Tianqi Wang et al., Oct. 2019 [5] | National School of Computer Science and Systems Analysis (ENSIAS), Mohammed V University Fenjiro Youssef et al., Oct. 2019 [3] | Graduate School of System Informatics, Kobe University Rousslan F. J. DOSSA et al., Sep. 2020 [10] | The Harker Upper School Heidi Lu and Olivia Xu, Mar. 2021 |
| Methods | Imitation and reinforcement learning, Airsims, ResNet-34, DDPG | DQfD, A2C model, Imitation and reinforcement learning | Imitation and reinforcement learning, DDPG, sensitivity test | **New** Imitation learning CNN, **optimized** reinforcement learning reward collection, DDPG, AirSim's API |
| Main Conclusions | combined IL and RL showed better performance as compared to both pure imitation learning and pure DDPG | resulting A2CfDoC model outperformed several existing algorithms in terms of speed and accuracy | proposed hybrid agent exhibits behavior similar to that of human experts | saved 80% of training time compared to pure reinforcement learning training |
| Advantages | drove the car in various weather and environments | surpassed expertise level using RL | hybrid agents surpassed that of the human expert in each scenario | used one human controlled car to collect sufficient training data to speed up RL training process |
| Limitations | only tested in simulated environment | only tested in simulated environment | only benefit from a larger population | only tested in smaller real world environment in a mini-scale car |

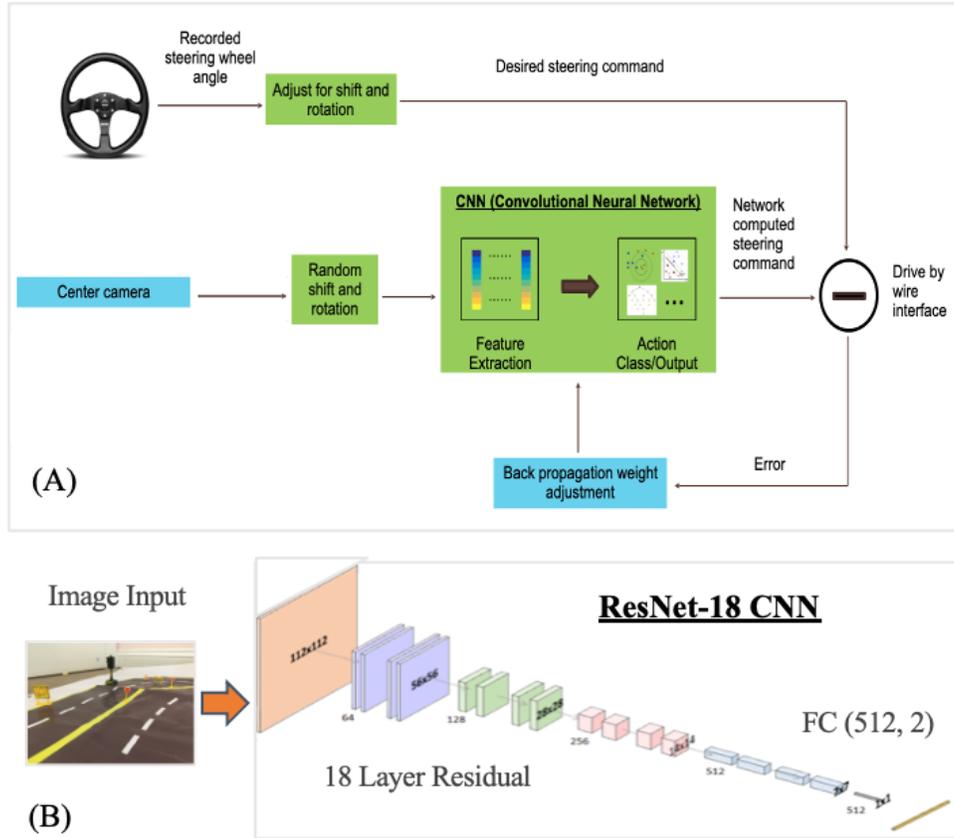

Figure 1 Design of the framework of supervised imitation learning developed to train the models as well as the integration of imitation learning and reinforcement learning. (A) shows a demonstration of human driving and how the data is used to pretrain a model for reinforcement learning and shaping the reward policy as well as new IL CNN developed, including specific parts like the camera and how the steering wheel inputs are adjusted for shift and rotation. (B) shows a more detailed view of the CNN, the ResNet-18 CNN network. It consists of 18 different layers, including a normalization layer, convolutional layers, and fully connected layers.

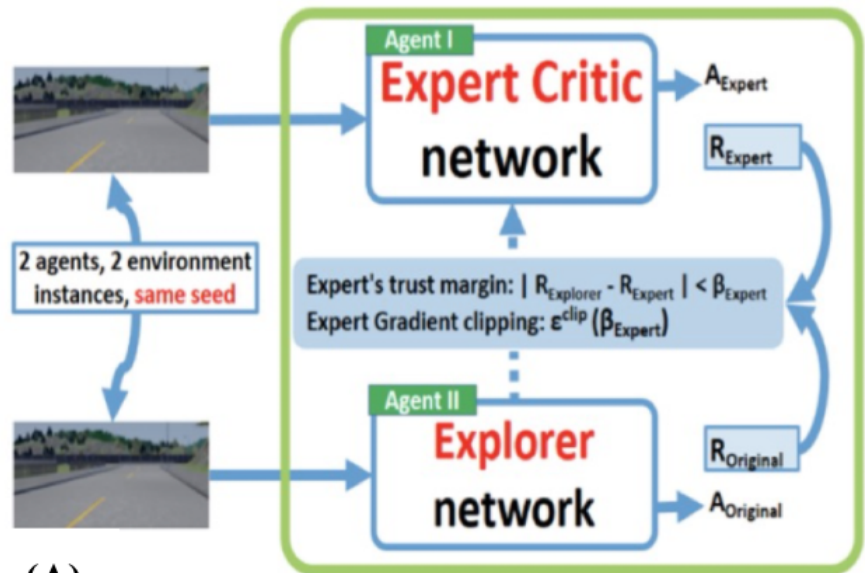

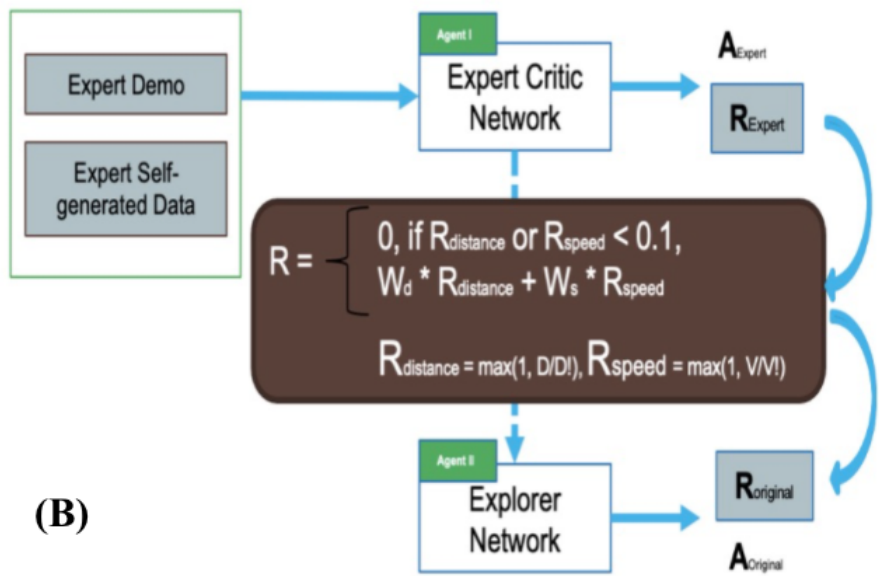

Figure 2 Two diagrams: previous existing architecture and the newly proposed architecture. (A) shows a demonstration of how driving using machine learning is better than human driving and how the existing architecture focuses on proving and making the rewards generated by reinforcement learning is better than the reward data generated by human expert driving data. (B) shows a diagram of how quickly machine learning can achieve a similar level to human expert driving, and how the new architecture works on making the reward generated by reinforcement learning better than the reward generated by human expert driving data in less training time.

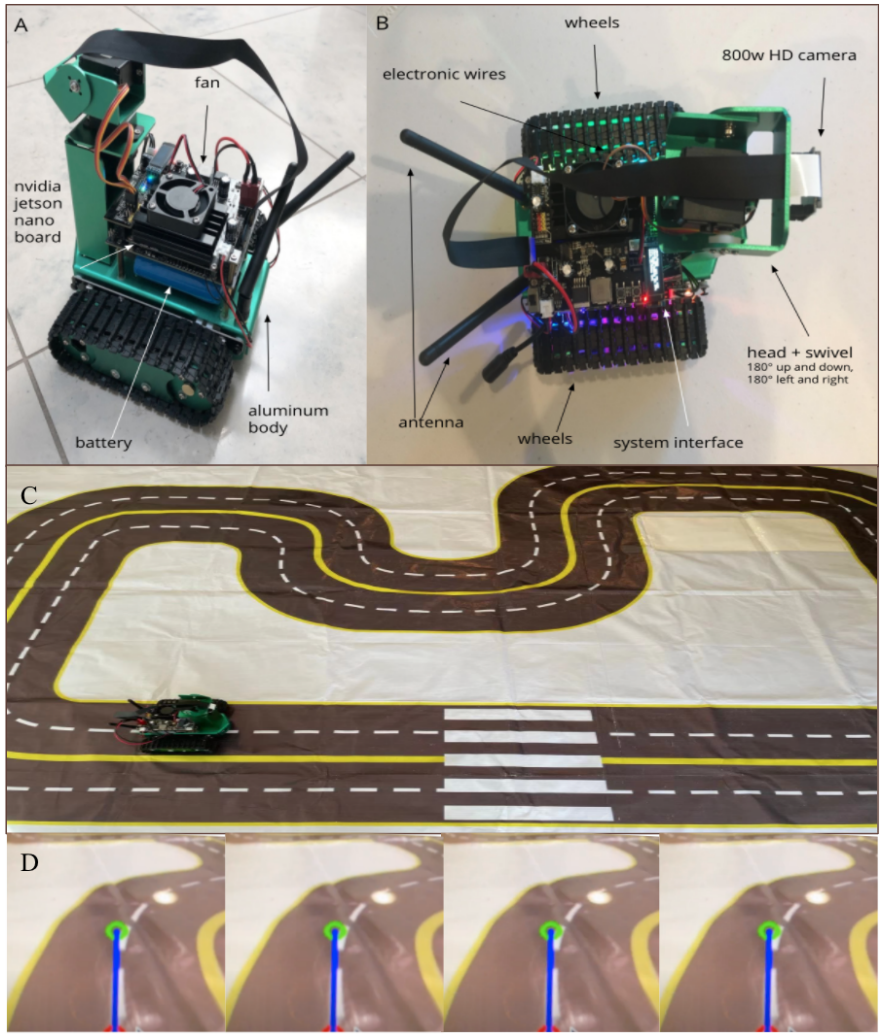

Figure 3 shows the testing procedures and hardware design, specifically the jetbot miniscale robot used. (A) and (B) shows the assembled car body as well as the integrated Raspberry Pi camera, steering gear, battery pack etc. (C) shows the real world environment set-up of a road map scenario (a 9 feet by 6 feet racing track) and testing the jetbot on it. (D) shows reward data collection situation where we're collecting some human driving data used for training.

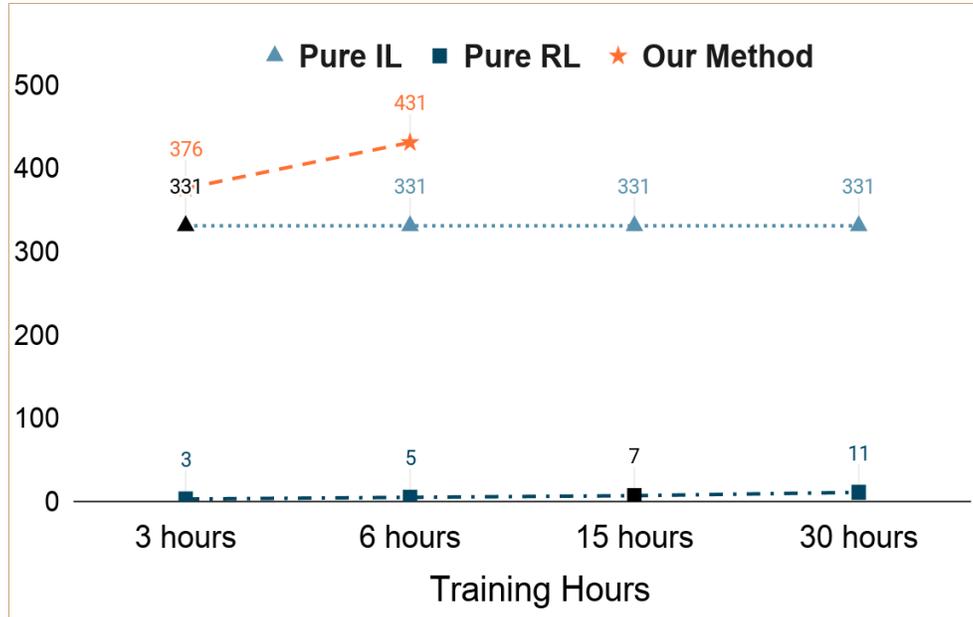

Figure 4 shows a graph of the Average Accumulated Reward where the orange star represents the method and the blue shapes represent pure IL and pure RL. As shown above, one can see the significant difference between the slow progress of the pure RL method and the straight line of pure IL. These numbers of the reward function are important to and could heavily affect the performance of the trained agent. For pure RL throughout a time period of 30 hours, the total reward accumulated was only 11.